\documentclass{article}

\usepackage{arxiv}

\usepackage[utf8]{inputenc} 
\usepackage[T1]{fontenc}    
\usepackage{hyperref}       
\usepackage{url}            
\usepackage{booktabs}       
\usepackage{amsfonts}       
\usepackage{nicefrac}       
\usepackage{microtype}      
\usepackage{lipsum}
\usepackage{graphicx}
\usepackage{subcaption}
\usepackage[numbers]{natbib}
\usepackage{array}
\usepackage{pdfpages}
\usepackage{authblk}

\graphicspath{ {./images/} }

\title{Evaluating LLMs on Generating Age-Appropriate Child-Like Conversations}

\author[1]{Syed Zohaib Hassan}
\author[1,2]{Pål Halvorsen}
\author[3,4]{Miriam S. Johnson}
\author[5,6]{Pierre Lison}

\affil[1]{Department of Holistic Systems, SimulaMet, Oslo, Norway}
\affil[2]{Department of Computer Science, Oslo Metropolitan University Oslo, Norway}
\affil[3]{Department of Behavioural Sciences, Oslo Metropolitan University, Oslo, Norway}
\affil[4]{Department of Psychology, Harvard University, Cambridge, MA, USA }
\affil[5]{Department of Informatics, University of Oslo, Oslo, Norway}
\affil[6]{Norwegian Computing Center, Oslo, Norway}

\affil[ ]{\texttt{syed@simula.no, paalh@simula.no, miriam.sinkerud-johnson@oslomet.no, plison@nr.no}}
 
\begin{document}
08:40\maketitle

\begin{abstract}

Large Language Models (LLMs), predominantly trained on adult conversational data, face significant challenges when generating authentic, child-like dialogue for specialized applications. We present a comparative study evaluating five different LLMs (GPT-4, RUTER-LLAMA-2-13b, GPTSW, NorMistral-7b, and NorBloom-7b) to generate age-appropriate Norwegian conversations for children aged 5 and 9 years. Through a blind evaluation by eleven education professionals using both real child interview data and LLM-generated text samples, we assessed authenticity and developmental appropriateness. Our results show that evaluators achieved strong inter-rater reliability (ICC=0.75) and demonstrated higher accuracy in age prediction for younger children (5-year-olds) compared to older children (9-year-olds). While GPT-4 and NorBloom-7b performed relatively well, most models generated language perceived as more linguistically advanced than the target age groups. These findings highlight critical data-related challenges in developing LLM systems for specialized applications involving children, particularly in low-resource languages where comprehensive age-appropriate lexical resources are scarce.

\end{abstract}

\keywords{Large Language Models (LLMs)\and Child Lexicon \and Low-Resource Languages  \and Age-of-Acquisition (AoA)}

\section{Introduction}

Large Language Models (LLMs) such as ChatGPT and GPT-4 have demonstrated remarkable capabilities in natural language generation across diverse domains~\cite{brown2020language,achiam2023gpt}. However, these models face significant challenges when applied to specialized applications requiring authentic child-like conversation generation. Most LLMs training datasets predominantly consist of adult-authored content from web crawls, books, and articles~\cite{dodge2021documenting,velu20245}, creating a fundamental gap in child-specific linguistic patterns and developmental markers necessary for generating authentic child-like text.

This limitation presents particular challenges for applications such as educational content development, and conversational agents designed for pediatric populations~\cite{druga2017hey,zhang2022storybuddy}. Research has shown that children's engagement with educational technology is significantly influenced by the developmental appropriateness of language used in interactive systems~\cite{kory2014storytelling}. The ability to generate developmentally appropriate language becomes critical in these contexts, where linguistic authenticity directly impacts effectiveness and appropriateness.

The challenge is amplified in low-resource languages such as Norwegian, where digital linguistic resources are substantially more limited compared to high-resource languages like English~\cite{joshi2020state}. While English benefits from extensive child language corpora and developmental databases~\cite{macwhinney2000childes,storkel2010online}, Norwegian lacks comparable resources for age-stratified lexical development, making it difficult to establish benchmarks for age-appropriate language generation.

This work addresses these challenges through a novel comparative methodology that systematically evaluates multiple LLMs' capability to generate age-appropriate Norwegian conversations. We focus specifically on two critical developmental stages (5 and 9 years old) that represent key transition points in early childhood (pre-school to school entry) and middle childhood (early primary school), covering the primary demographic for educational avatar applications. Our study contributes both methodological insights for evaluating specialized language generation tasks and empirical findings about current LLMs' limitations in low-resource, domain-specific applications.


\section{Related Work}

Child language acquisition follows well-documented developmental patterns, with vocabulary, syntax, and pragmatic skills evolving systematically across age groups~\cite{bloom2002children}. Computational approaches to modeling child language development have traditionally focused on acquisition patterns~\cite{macwhinney2010computational} and developmental lexicons~\cite{fenson2007macarthur}. Age-of-Acquisition (AoA) ratings have proven crucial for understanding lexical development, with earlier-acquired words typically being shorter, more frequent, and less morphologically complex~\cite{kuperman2012age}. While most of these research works focus on high-resource languages like English, similar studies exist for Norwegian \cite{jensen2024statistics,kristoffersen2012utvikling,simonsen2014norwegian}. However, the intersection of LLMs and child-like conversation generation remains largely underexplored, particularly for low and mid-resource languages. These psycholinguistic foundations provide essential benchmarks for evaluating computational LLMs' ability to generate age-appropriate language. 

Recent studies have highlighted critical limitations in LLMs' ability to generate age-appropriate content for children. Bhandari et al.~\cite{bhandari2023trustworthiness} assessed the trustworthiness of children's stories generated by LLMs, revealing that while generated stories may resemble real ones in themes and patterns, they often lack the nuances of genuine children's literature and may include inappropriate content. Their work demonstrates that LLMs are not yet suitable for generating quality children's content without careful oversight.

Valentini et al.~\cite{valentini2023automatic} focused on lexical and readability levels in LLM-generated children's stories, finding that while LLMs can generate relevant content, they frequently fail to adjust vocabulary to meet comprehension levels appropriate for younger children. This research underscores the critical gap in LLMs' ability to generate truly age-appropriate content. The challenges of LLM adaptation become particularly acute in low-resource language contexts, where training data scarcity compounds domain-specific limitations~\cite{hedderich2020survey}.

While existing research identifies problems with LLM-generated child content, there is limited systematic comparative evaluation across multiple models, and methodological frameworks for assessing age-appropriateness specifically in conversation generation remain unexplored. Furthermore, most existing work focuses on English, leaving significant gaps in understanding these challenges across different language contexts, particularly for low-resource languages like Norwegian, where specialized linguistic resources are limited.

\section{Methodology}

We designed a comprehensive comparative evaluation study to assess multiple LLMs' ability to generate age-appropriate child-like conversations in Norwegian. Our methodology combines expert human evaluation with computational linguistic analysis to provide both qualitative and quantitative assessments of model performance.


\subsection{Model Selection and Description}

We evaluated five diverse LLMs representing different architectural approaches and training paradigms, selected to include both Norwegian-specific and state-of-the-art multilingual models:
\begin{itemize}
    \item RUTER-LLAMA-2-13b~\cite{llama2_13b_chat}: A 13-billion parameter model built upon the LLaMA-2~\cite{touvron2023llama} foundation and adapted for Norwegian through fine-tuning. This model demonstrates the approach of leveraging established multilingual architectures for low-resource language applications.
    \item GPTSW~\cite{gptsw3}: A 6.7-billion parameter language model is pre-trained for Scandinavian language processing (Swedish, Danish, Norwegian, and Icelandic) to capture linguistic characteristics and cultural context, specific to Scandinavian countries.
    \item GPT-4~\cite{achiam2023gpt}: OpenAI's flagship large language model, at the time, accessed through API integration. We included this model as a benchmark representing current state-of-the-art multilingual generation capabilities, despite its general-purpose rather than Norwegian-specific training.
    \item NorMistral-7b~\cite{NorMistral7b}: A 7-billion parameter  adaptation of the Mistral framework~\cite{jiang2023mistral7b} instruction-tuned for Norwegian language tasks. This model represents contemporary approaches to efficient language modeling specifically designed for resource-constrained linguistic environments.
    \item NorBloom-7b~\cite{NorBloom7b}: A 7-billion parameter Norwegian-specialized variant of the BLOOM architecture~\cite{workshop2022bloom}, specifically instruction-tuned for handling Norwegian language.
\end{itemize}

This model selection strategy allows us to compare Norwegian-specific open-source models against state-of-the-art multilingual systems, providing insights into the relative merits of specialized versus general-purpose approaches for low-resource language applications.

\subsection{Real Child Interview Data Collection}

We collected authentic interview data from children under ethical approval from the Norwegian Agency for Shared Services in Education and Research (SIKT, application number 385370).

The data collection involved face-to-face structured interviews with children aged 4-12 years, conducted using a standardized script consisting of 10 questions covering neutral topics related to their everyday life, including their favorite food, their favorite toys and activities, and everyday activities. To facilitate further elaboration, children's initial responses were followed up with open-ended questions. This standardized approach ensured consistency across all interviews while maintaining natural conversational flow. Each interview lasted approximately 15 minutes. All interviews were audio-recorded during the sessions for subsequent processing. Following each interview, the audio recordings were transcribed verbatim and fully anonymized to remove any identifying information. After transcription, all original audio recordings were permanently deleted to ensure participant privacy and data security. All participants received a gift card as compensation for their participation.

From the collected data comprising 10 interviews distributed across ages (two interviews each for ages 5, 6, 7, and 8, and one interview each for ages 9 and 10), we selected representative samples for our target age groups: one 5-year-old interview and one 10-year-old interview. These samples were chosen based on typical linguistic development markers for their respective age groups and conversation length suitable for expert evaluation. The selected interviews represented both male and female participants to ensure gender representation in our authentic child language samples.


\subsection{LLM Text Generation Protocol}\label{llm-generation}

Each LLM generated conversational text samples using identical prompts based on the same standardized script of 10 questions used in the real child interviews. This approach ensured direct comparability between authentic child responses and LLM-generated content by maintaining consistent conversational contexts and topics across all samples. The prompts were designed to elicit age-appropriate dialogue for 5-year-old and 9-year-old children while simulating natural conversation scenarios.
For each target age group, we generated samples representing both male and female child personas across all models, resulting in multiple candidate texts per model per age group. From this larger set of generated content, we randomly selected one sample per model per age group for inclusion in our evaluation study, with gender selection occurring randomly to avoid systematic bias. This process resulted in 10 LLM-generated texts total (5 models × 2 age groups) that maintained gender diversity while ensuring unbiased selection.
Generation parameters were standardized across models where possible to ensure fair comparison, with temperature settings optimized for natural conversation flow while maintaining consistency across responses.

\subsection{Norwegian Linguistic Resources}

We supplemented the subjective user study with an objective computational analysis of the linguistic characteristics in the LLM-generated conversation texts. To support computational analysis, we compiled a comprehensive Norwegian child language development dataset from openly available linguistic resources. Table~\ref{tab:norwegian_language_data} presents an overview of various linguistic datasets used in our evaluation of child language development in Norwegian. The compiled dataset includes detailed linguistic features such as subjective age-of-acquisition (AoA) ratings, usage frequency, and structural characteristics like the number of letters, sounds, and syllables for each word. This collection comprises ten distinct datasets that collectively contain $2,162$ words, of which $1,963$ (approximately 91\%) have been annotated with AoA information. The datasets originate from different research initiatives focused on Norwegian language acquisition and development.

\begin{table}[t]
\footnotesize
\centering
\caption{ Language Datasets with child lexicon development information}
\begin{tabular}{lrr}
\hline
\textbf{Dataset} & \textbf{Number of words} & \textbf{Number with AoA} \\
\hline
CAT~\cite{jensen2024statistics} & 104 & 73 \\
CDI-I~\cite{kristoffersen2012utvikling} & 562 & 546 \\
CDI-II~\cite{simonsen2014norwegian} & 478 & 464 \\
CDI-III~\cite{holm2023norwegian} & 49 & 2 \\
CLT~\cite{haman2015designing} & 127 & 127 \\
NUMA~\cite{lind2015norwegian} & 214 & 190 \\
PALPA~\cite{kay2009psykolingvistisk} & 392 & 392 \\
The-Association-Test~\cite{boyum2016new} & 70 & 3 \\
The-Past-Tense~\cite{ragnarsdottir1999acquisition} & 58 & 58 \\
VAST~\cite{lind2007verb} & 108 & 108 \\ 
\hline
\textbf{Grand Total} & \textbf{2162} & \textbf{1963} \\
\hline
\end{tabular}
\label{tab:norwegian_language_data}
\end{table}

\begin{figure*}[ht]
    \centering
    \includegraphics[width=0.7\linewidth]{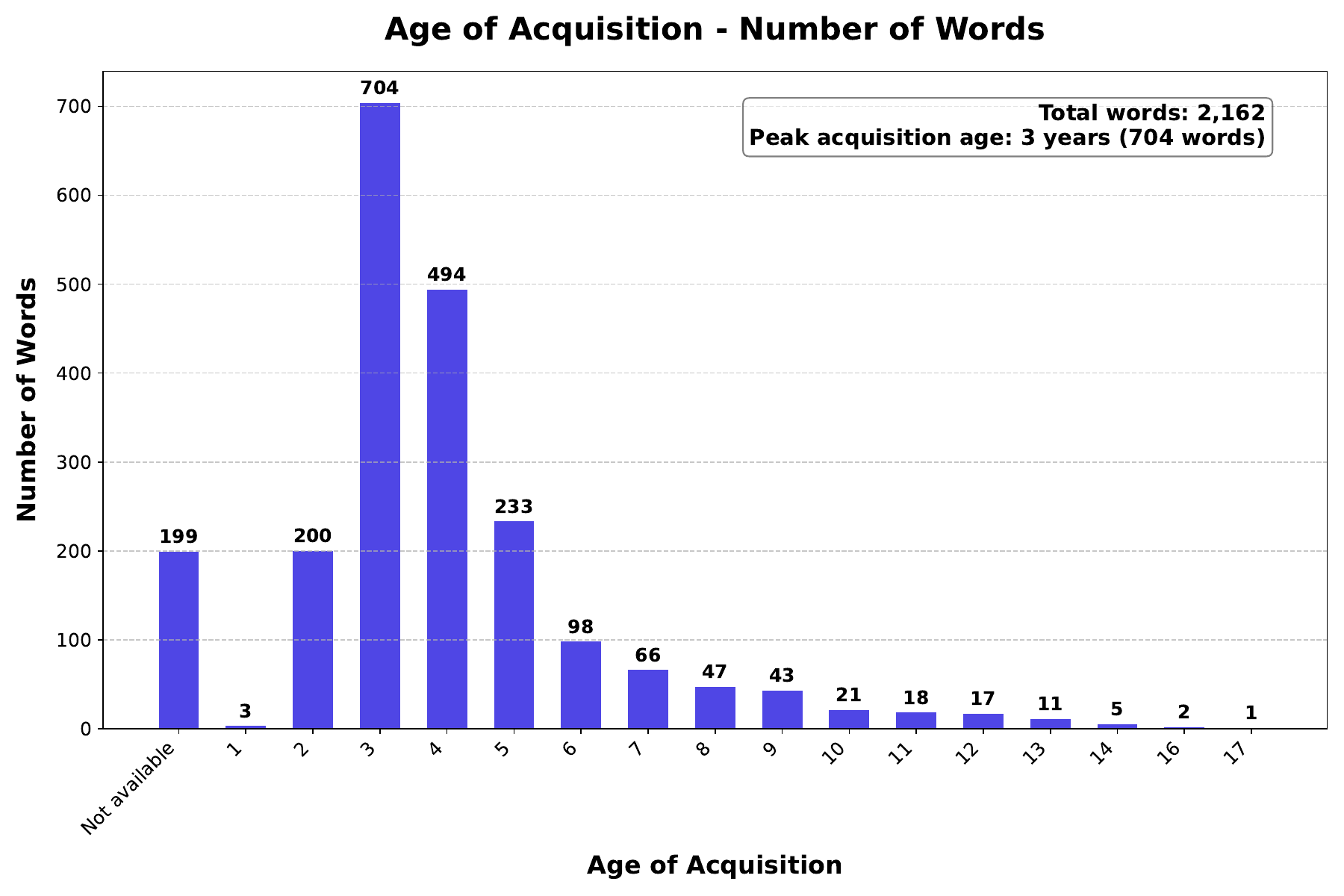}
    \caption{Age prediction statistics by age for consolidated dataset}
    \label{fig:aoa-stats}
\end{figure*}

Following data preprocessing, which involved duplication removal and the incorporation of inflectional forms, our final dataset comprised $1,813$ Norwegian words. This dataset consists of $1,095$ base words and $718$ inflectional forms. The distribution of word classes among the base words revealed a predominance of nouns (n = $713$), followed by verbs (n = $307$) and adjectives (n = $75$). A significant limitation we encountered is that 83\% ($1,580$ words including inflections) of the words with AoA scores in our compiled dataset are associated with children under age 5. Figure~\ref{fig:aoa-stats} shows a heavy skew toward early childhood vocabulary in the consolidated dataset. This created analytical challenges when attempting to evaluate the age appropriateness of LLM-generated texts for the 9-year-old samples in our study. This limitation highlights the broader challenge of developing comprehensive linguistic resources for low-resource languages like Norwegian.

\subsection{Evaluation Methodology}

Our evaluation framework combined expert human assessment with computational linguistic analysis to provide a comprehensive evaluation of age-appropriate language generation. This integrated methodology enables both qualitative assessment of developmental authenticity and quantitative analysis of linguistic characteristics.

\subsubsection{Evaluation Protocol}

We recruited ten education professionals to conduct blind evaluations of all text samples. The evaluator group comprised four educational-psychological counselors (\textit{rådgivere} from \textit{Pedagogisk-Psykologisk Tjeneste} - PPT), three teachers with classroom experience (including \textit{adjunkter}, \textit{lektorer}, and \textit{lærere}), one counselor and teacher,  and two child welfare professionals (one current specialist and one former specialist who is now a parent). Eight evaluators had ten or more years of professional experience working with children, while two had 5-7 years of experience. The evaluators' professional experience collectively spanned the full age range from 0 to 18 years. None of the evaluators had previously participated in formal studies related to child language development, ensuring their assessments reflected professional expertise rather than specialized research training.

Educational-psychological counselors, trained specifically in assessing children's learning and developmental patterns, add further specialized expertise to developmental assessment. Together, these education professionals bring complementary observational and assessment skills that position them well to evaluate child language development stages for our study.

Teachers bring expertise through systematic classroom-based observation skills. Research indicates that teachers of early years children can make judgments about children's language skills with relatively high sensitivity (92\%) and specificity (85.7\%) when compared with formal testing outcomes~\cite{williams2006teacher}. The National Association for the Education of Young Children (NAEYC) emphasizes that observing and documenting children's development are essential professional competencies~\cite{naeyc2023}. Educational-psychological counselors, trained specifically in assessing children's learning and developmental patterns, add further specialized expertise to developmental assessment.

Each evaluator independently assessed 12 text samples (six per age group, including one real interview and five LLM-generated texts per group), detailed in Table~\ref{app:sample-details} in appendix. For each sample, evaluators were asked to: Predict the age of the child represented in the text, and provide written justification for their age assessment. Evaluators were not informed which samples were real versus LLM-generated, nor were they told the intended target ages, ensuring unbiased assessment based solely on linguistic content. This blind evaluation design enables direct comparison of expert perceptions across different text sources and generation methods while leveraging the naturalistic expertise that education professionals develop through extensive, regular observation of children's language development. The user study form is attached in Appendix~\ref{app:child-lexical}.

\subsubsection{Computational Linguistic Analysis}

To complement expert evaluation with objective measures, we conducted systematic computational analysis of all text samples using a compiled Norwegian child language development dataset. We compiled this comprehensive dataset from ten established linguistic resources (Table~\ref{tab:norwegian_language_data}). The compiled dataset allowed us to conduct computational analysis, using linguistic development characteristics, including: (i) Average AoA score; (ii) Average frequency per million words; (iii) Structural Characteristics (Average number of letters, syllables, and phonemes per word), and (iv) response length (Total word count).

This multi-faceted approach enables correlation between expert perceptions and objective linguistic measures, providing insights into which computational features best predict perceived age-appropriateness and validating the relationship between subjective expert assessment and quantifiable linguistic characteristics.

\section{Results}

We present our findings in four main areas: expert age prediction performance, inter-rater reliability, statistical analysis of prediction patterns, and computational linguistic analysis. These results provide both quantitative and qualitative insights into how well different LLMs generate age-appropriate Norwegian child-like conversations, as assessed by education professionals and objective linguistic metrics.

\subsection{Education Professional Age Prediction Performance}

Tables~\ref{tab:text-statistics-age5} and~\ref{tab:text-statistics-age9-10} present comprehensive patterns in education professionals' ability to accurately predict children's ages based on text samples across both age groups and text sources.

\subsubsection{Age Prediction Accuracy and Model Comparison}

Table~\ref{tab:text-statistics-age5} shows that for the 5-year-old category, all LLM-generated texts except NorMistral-7b were consistently overestimated in age, with mean predicted ages ranging from $5.65$ to $5.90$ years. Notably, the real interview for the 5-year-old was perfectly estimated with a mean prediction of exactly 5 years and the lowest standard deviation ($1.03$), indicating strong consensus among evaluators. Conversely, NorMistral-7b's generated text was significantly underestimated at $3.85$ years.

Table~\ref{tab:text-statistics-age9-10} shows that for the older age group, we observe greater variability. The real interview (from a 9-year-old child) was overestimated at $12.05$ years with the lowest standard deviation ($0.69$), suggesting high agreement among participants. RUTER-LLAMA-2-13b showed the largest discrepancy, with its 9-year-old text being estimated at nearly 14 years, an overestimation of almost 5 years. NorMistral-7b and NorBloom-7b performed most accurately for this age group, with mean errors of just $0.45$ and $0.60$ years, respectively.

\begin{table}[t]
\footnotesize
\centering
\caption{Summary Statistics for Age 5 Texts}
\label{tab:text-statistics-age5}
\begin{tabular}{lccccl}
\hline
\textbf{Source} & \textbf{Actual Age} & \textbf{Mean Predicted Age} & \textbf{Std Predicted Age} & \textbf{Mean Error} & \textbf{Text Source} \\
\hline
Real & 5 & 5.00 & 1.03 & 0.00 & Real Interview \\
LLM & 5  & 5.75 & 1.43 & 0.75 & RUTER-LLAMA-2-13b \\
LLM & 5  & 5.90 & 2.32 & 0.90 & GPTSW \\
LLM & 5  & 5.65 & 1.86 & 0.65 & GPT-4 \\
LLM & 5  & 3.85 & 1.42 & 1.15 & NorMistral-7b \\
LLM & 5  & 5.85 & 1.76 & 0.85 & NorBloom-7b \\
\hline
\end{tabular}
\footnotesize
\end{table}

\begin{table}[t]
\footnotesize
\centering
\caption{Summary Statistics for Age 9 Texts}
\label{tab:text-statistics-age9-10}
\begin{tabular}{lccccl}
\hline
\textbf{Source} & \textbf{Actual Age} & \textbf{Mean Predicted Age} & \textbf{Std Predicted Age} & \textbf{Mean Error} & \textbf{Text Source} \\
\hline
Real & 9 & 12.05 & 0.69 & 3.05 & Real Interview \\
LLM & 9 & 13.90 & 2.23 & 4.90 & RUTER-LLAMA-2-13b \\
LLM & 9 & 10.45 & 1.95 & 1.45 & GPT-4 \\
LLM & 9 & 8.40 & 2.63 & 0.60 & NorBloom-7b \\
LLM & 9 & 11.70 & 1.87 & 2.70 & GPTSW \\
LLM & 9 & 8.55 & 1.89 & 0.45 & NorMistral-7b \\
\hline
\end{tabular}
\footnotesize
\end{table}

\begin{figure*}[ht]
    \centering
    \includegraphics[width=\linewidth]{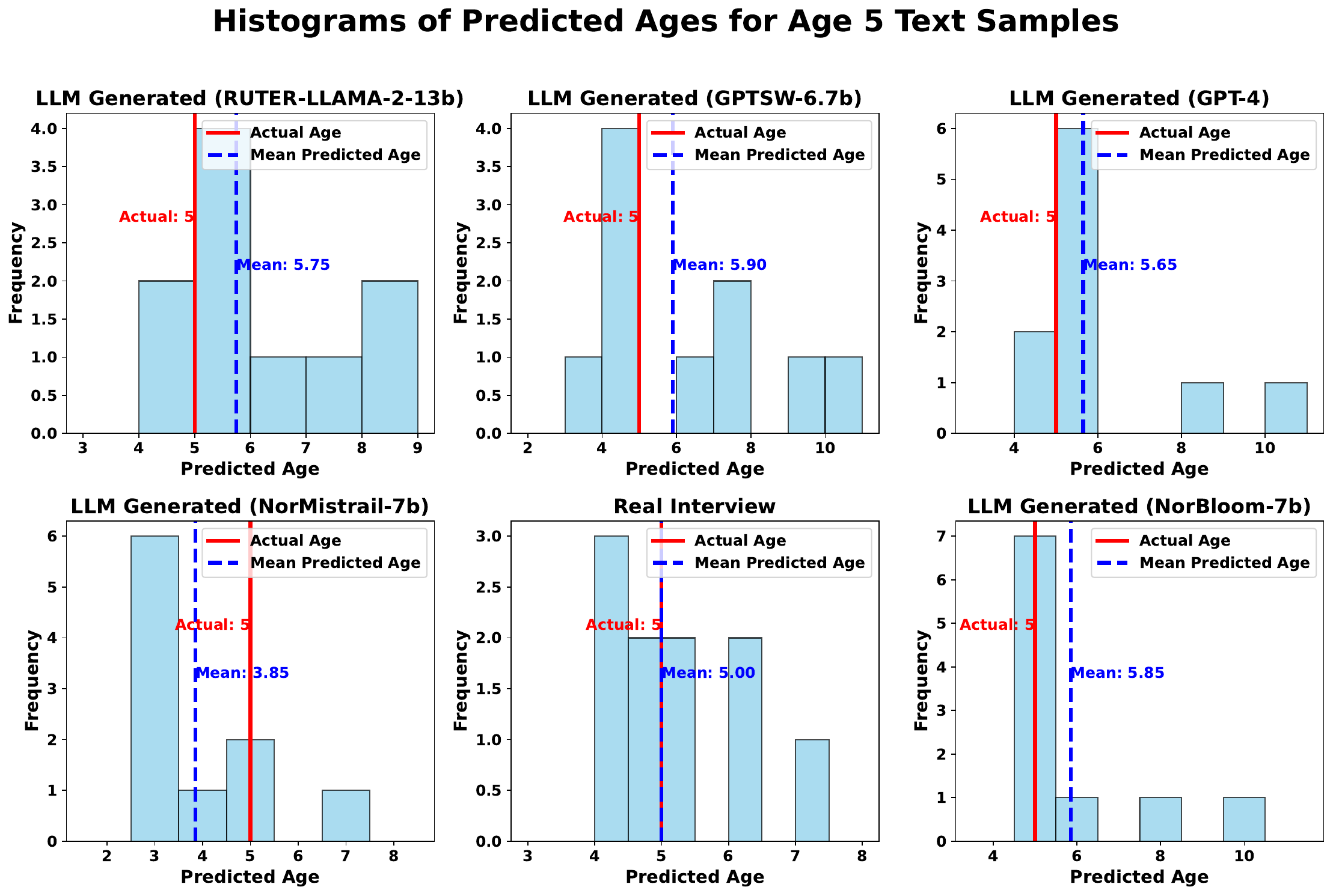}
    \caption{Age prediction histograms for age 5}
    \label{fig:histograms5}
\end{figure*}

\begin{figure*}[ht]
    \centering
    \includegraphics[width=\linewidth]{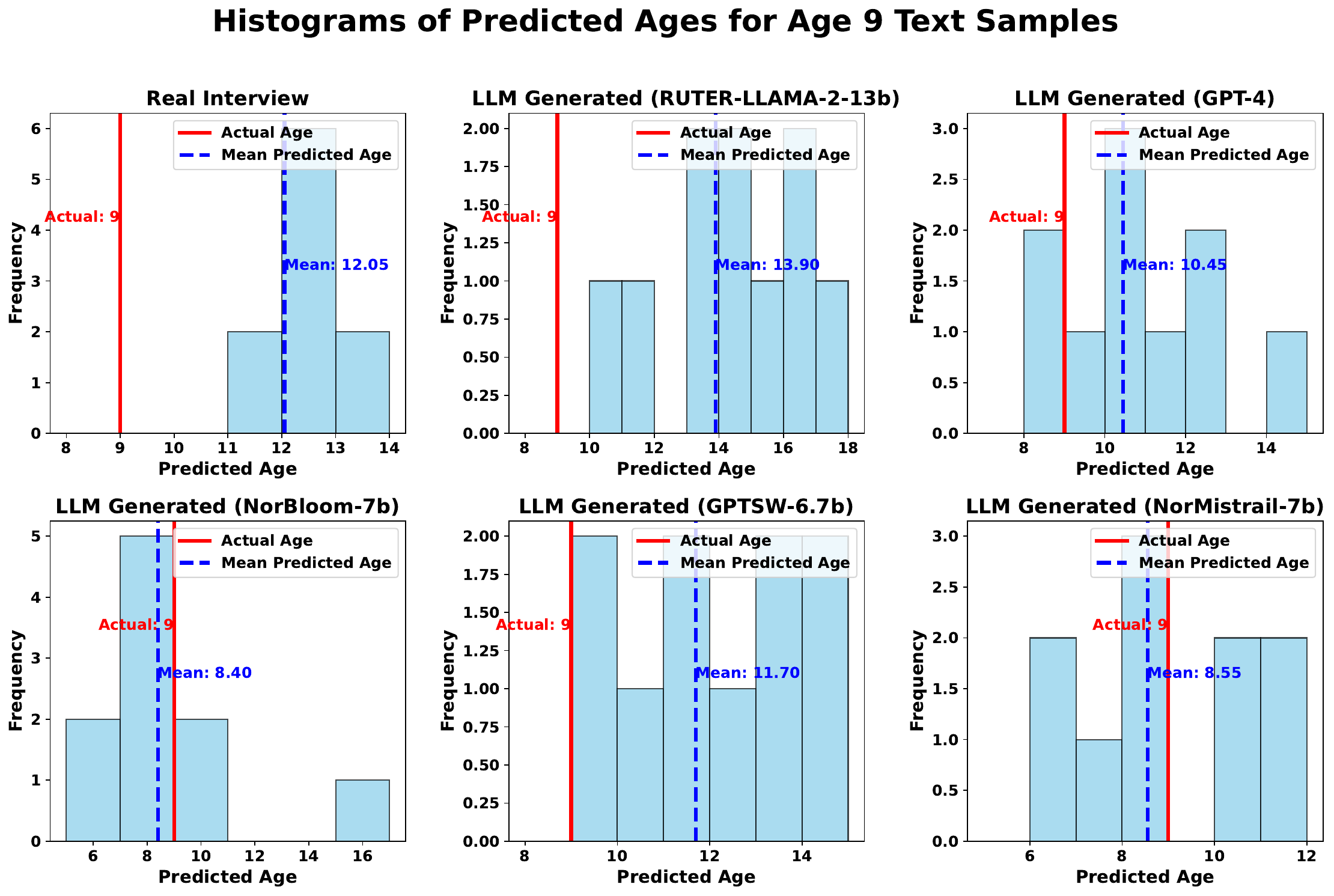}
    \caption{Age prediction histograms for age 9}
    \label{fig:histograms9}
\end{figure*}

The distribution patterns shown in Figures \ref{fig:histograms5} and \ref{fig:histograms9} further illustrate these age-related differences, with 5-year-old predictions showing tighter clustering around target ages compared to more dispersed predictions for older children. The results reveal the distribution of these age predictions across all 12 text samples. Among LLM-generated texts, GPT-4 and NorBloom-7b produced texts that resulted in the most accurate age predictions overall, with mean absolute errors (MAE) of $1.40$ and $1.48$ years, respectively. NorMistrail-7b also performed well with an MAE of $1.60$ years. In contrast, RUTER-LLAMA-2-13b showed the largest discrepancy between actual and predicted ages (MAE: $2.98$ years).

Performance varied by age group. For 5-year-old texts, GPT-4 and NorBloom-7b both achieved the lowest prediction errors (MAE: $0.95$ years), while GPTSW-6.7b showed the highest error (MAE: $1.90$ years). The difference between models was more pronounced for 9-year-old texts, where GPT-4 maintained
the best performance (MAE: $1.85$ years) but RUTER-LLAMA-2-13b showed substantially larger errors (MAE: $4.90$ years), nearly 2.6 times higher than the best-performing model.

\subsubsection{Age Group Differences in Prediction Accuracy}
Statistical analysis revealed significantly different prediction accuracy patterns between age groups. Evaluators were generally more accurate in predicting the language of 5-year-olds (MAE = $1.20$ years) compared to 9-year-olds (MAE = $2.69$ years). This suggests that the linguistic markers of early childhood may be more distinct and recognizable than those of middle childhood. Most LLMs tended to produce language that appeared slightly more advanced than the target age, with 7 out of 10 LLM-generated texts having mean predicted ages higher than their actual ages. The exceptions were the texts generated by NorMistral-7b, which produced the language that was perceived as slightly less mature than the target ages.

\begin{table}[htbp]
\centering
\footnotesize
\caption{Intraclass Correlation Coefficient (ICC) Results for Rater Agreement}
\label{tab:icc-results}
\begin{tabular}{ll}
\hline
\textbf{Measure} & \textbf{Value} \\
\hline
Model & Two-way random effects \\
Type & Absolute agreement \\
Number of subjects (texts) & 12 \\
Number of raters & 10 \\
ICC(2,1) & 0.754 \\                          
F-statistic & F(11, 99) = 32.67 \\ 
p-value & p $<$ 0.001 \\
95\% Confidence Interval & 0.733 -- 0.776 \\
\hline
\end{tabular}
\end{table}

\subsection{Inter-Rater Reliability Analysis}

The inter-rater reliability for age predictions in our lexical study was assessed using two-way random effects Inter-Rater Correlation Coefficient (ICC) for absolute agreement, as shown in Table \ref{tab:icc-results}. The ICC was calculated using classical ANOVA formulas~\cite{shrout1979intraclass}. The analysis yielded an ICC(2,1) value of $0.754$, with a $95\%$ confidence interval ranging from $0.733$ to $0.776$. According to established guidelines for interpreting ICC values~\cite{koo2016guideline}, our result indicates good reliability 
among the education professionals who participated in this study.

The ICC value of $0.754$ suggests that $75\%$ of the variance in age predictions can be attributed to true differences between the text samples, while the remaining $25\%$ represents measurement error arising from differences in how individual raters perceived and evaluated the texts. This level of agreement is interesting, considering the subjective nature of age estimation based solely on linguistic features. The strong agreement among raters was further supported by the highly significant F-test result (F(11, 99) = $32.67$, p $<$ $0.001$), confirming that the observed agreement did not occur by chance. The narrow confidence interval (0.733 to 0.776) indicates good precision in the reliability estimate, with both bounds falling within the "good" reliability range.

Text presentation order was fixed across all raters, which prevents separation of true inter-rater agreement from potential order effects e.g., fatigue or practice, or familiarity effects). Future studies should randomize text order per rater to avoid this confound.

\subsection{Statistical Analysis of Prediction Patterns}

To further understand the patterns observed in expert age predictions, we conducted statistical analyses examining differences between text sources and age groups. These analyses provide quantitative support for the qualitative patterns observed in the descriptive statistics and help identify systematic biases in both LLM text generation and education professionals' assessment.

\subsubsection{ Paired T-Test Results}
We also conducted paired t-tests to examine the differences in age prediction accuracy between real and LLM-generated texts, as well as between different age groups. The within-subject design allowed us to account for individual participant variability while comparing prediction performance across conditions.

Our analysis revealed no significant difference in prediction errors between authentic interviews (M = $1.93$, SD = $0.50$) and LLM-generated texts (M = $1.95$, SD = $0.70$), t(9) = $-0.114$, p = $0.912$. This finding suggests that evaluators' overall accuracy in age prediction was comparable regardless of text source, indicating that some LLMs can generate text that achieves similar levels of perceived authenticity as real child language.

The comparison between age groups revealed significant differences in prediction accuracy. Texts from 5-year-olds (M = $1.20$, SD = $0.79$)  were predicted significantly 
more accurately than those from 9-year-olds (M = $2.69$,
SD = $0.73$), t(9) = $-5.352$, p < $0.001$. This statistically significant difference confirms that expert assessment of younger children's language is substantially more accurate than assessment of older children's language, suggesting that linguistic markers of early childhood are more distinct and recognizable than those of middle childhood.

\subsubsection{Error Pattern Analysis}

Figures \ref{fig:histograms5} and \ref{fig:histograms9} reveals systematic tendencies in evaluators' assessments. Most LLM-generated texts (7 out of 10) resulted in age overestimation, with experts perceiving the language as more linguistically advanced than the target age. The exceptions were both texts generated by NorMistral-7b, which consistently produced language perceived as slightly less mature than the target ages.
The consistently higher standard deviations for the 9-year-old category indicate greater difficulty in precisely estimating age for older children, possibly reflecting wider individual variability in language development during middle childhood compared to early childhood.

\subsection{Computational Linguistic Analysis}

We conducted computational analysis of the linguistic characteristics present in both real and LLM-generated texts. Table \ref{tab:llm_stats} presents comprehensive metrics including response length, average Age-of-Acquisition (AoA) scores, word frequency, and structural characteristics for each text sample.

A notable limitation in our analysis stems from the AoA data discussed earlier. With 83\% of words in our Norwegian linguistic database having AoA ratings for children under 5 years, the differentiation between text generated for 5-year-olds versus 9-year-olds becomes less distinct when relying solely on AoA metrics. This is reflected in the minimal differences in average AoA scores between the two age groups across all LLMs and real data, with values ranging narrowly from $3.39$ to $3.60$ for 5-year-olds and $3.60$ to $3.94$ for 9-year-olds. While there is a slight increase in AoA for the older age group, this increase is insufficiently distinctive to serve as a reliable indicator of age-appropriate language.

The word frequency data provides some interesting patterns, particularly for real-life data, NorMistral-7b and NorBloom-7b, which show substantially higher average frequency per million words ($7725.94$, $8261.09$, and $7523.32$, respectively) in their 9-year-old text samples compared to their 5-year-old samples. Since higher frequency words are typically acquired at younger ages, this counterintuitive pattern might help explain why NorMistral-7b's 9-year-old text was underestimated (mean predicted age: 8.55) in the subjective evaluation.

Response length varies dramatically across LLMs, with GPT-4 producing the most verbose responses for both age groups ($67.60$ and $81.20$ words on average), while NorBloom-7b and NorMistral-7b generated much shorter responses (as low as $8.15$ words for NorMistral-7b's 5-year-old responses). This variation in verbosity does not appear to correlate directly with accuracy in age prediction, suggesting that content and style may be more influential than length in conveying age-appropriate speech.

The structural characteristics of real interviews (average number of letters, syllables, and sounds) fall within similar ranges as LLM-generated text, suggesting that these high-level metrics alone may not capture the qualitative differences that human experts readily detect in authentic child speech. 

The structural characteristics show subtle variations across LLMs and age groups, with no clear pattern distinguishing the more successful models identified in the subjective evaluations. This further underscores the challenge of using purely objective linguistic metrics for evaluating age-appropriate language generation in a low-resource language context, particularly with limited reference lexicon data.

\begin{table}[t]
\footnotesize
\centering
\caption{Characteristics of LLM-generated text by age}
\begin{tabular}{lcccccccc}
\hline
Age & LLM & \begin{tabular}[c]{@{}c@{}}Response\\ Length\end{tabular} & Avg AoA & \begin{tabular}[c]{@{}c@{}}Avg Freq\\ per Million\end{tabular} & \begin{tabular}[c]{@{}c@{}}Avg\\ Letters\end{tabular} & \begin{tabular}[c]{@{}c@{}}Avg\\ Syllables\end{tabular} & \begin{tabular}[c]{@{}c@{}}Avg\\ Sounds\end{tabular} \\
\hline
Five & GPT-4 & 67.60 & 3.49 & 5252.66 & 4.02 & 1.68 & 3.67 \\
Five & GPTSW & 26.75 & 3.47 & 3266.45 & 4.70 & 1.90 & 4.14 \\
Five & NorBloom-7b & 10.65 & 3.60 & 5016.33 & 4.04 & 1.77 & 3.77 \\
Five & NorMistral-7b & 8.15 & 3.45 & 4045.33 & 4.32 & 1.73 & 3.74 \\
Five & RUTER-LLAMA-2-13b & 46.75 & 3.39 & 4029.10 & 4.10 & 1.63 & 3.59 \\
Five & Real & 33.2 & 3.53 & 3147.34 & 4.4 & 1.74 &	3.91\\
\hline
Nine & GPT-4 & 81.20 & 3.70 & 5003.94 & 4.26 & 1.65 & 3.72 \\
Nine & GPTSW & 21.10 & 3.81 & 3033.87 & 4.87 & 1.87 & 4.13 \\
Nine & NorBloom-7b & 8.80 & 3.77 & 7523.32 & 3.97 & 1.67 & 3.57 \\
Nine & NorMistral-7b & 15.25 & 3.94 & 8261.09 & 3.98 & 1.83 & 3.86 \\
Nine & RUTER-LLAMA-2-13b & 50.10 & 3.60 & 4190.94 & 4.10 & 1.57 & 3.57 \\
Nine & Real & 31.8 & 3.86 & 7725.94 & 4.02  &1.69 & 3.77\\
\hline
\end{tabular}
\footnotesize
\label{tab:llm_stats}
\end{table}

\section{Discussion}

Our comprehensive evaluation of five LLMs' ability to generate age-appropriate Norwegian child-like conversations reveals both promising capabilities and significant limitations that have important implications for language generation applications involving children.

\subsection{Model Performance and Age-Related Challenges}

Our findings reveal several critical insights for the development and application of LLMs in generating child-like language for specialized applications. The differences across LLMs highlight that model size and general capabilities do not necessarily translate to better performance in specialized tasks requiring developmental linguistic knowledge. Notably, smaller Norwegian models pre-trained from scratch (NorBloom-7b and NorMistral-7b) showed relatively better performance compared to a larger fine-tuned multilingual model (RUTER-LLAMA-2-13b) in our evaluation. This suggests that pre-training on language-specific data may offer advantages over fine-tuning larger multilingual models. However, these findings require cautious interpretation given our limited sample size: two real interviews and ten LLM-generated texts. More robust conclusions would require additional generations per model and text-instance level analysis using mixed-effects models. GPT-4's relatively strong performance despite not being specifically trained on Norwegian demonstrates the potential value of leveraging large multilingual models with sophisticated prompt engineering strategies. This suggests a practical approach for low-resource languages: careful adaptation of existing high-capacity multilingual systems may prove more effective and resource-efficient than training entirely new models from scratch. 

Most LLMs generate language perceived as more linguistically advanced than target ages, pointing to a fundamental challenge in current training paradigms. This systematic bias toward linguistic sophistication likely reflects the adult-centric nature of most training corpora, where even content for children is typically authored by adults using more complex linguistic structures than children naturally produce. The greater difficulty in generating appropriate language for 9-year-olds compared to 5-year-olds reflects increasing complexity and individual variability in language development as children mature. This pattern suggests age-specific prompt engineering strategies may be necessary for different developmental stages.

\subsection{Methodological Insights and Evaluation Limitations}

The strong inter-rater reliability (ICC = 0.75) demonstrates that education professionals can consistently identify age-related linguistic features, even without contextual information about text source or intended target age. This validates using education professionals' evaluation as a reliable methodology for assessing age-appropriate language generation. Higher accuracy in predictions for younger children suggests early childhood linguistic markers are more distinctive and recognizable than those of middle childhood. This may reflect both the more dramatic developmental changes in early language acquisition and greater consensus regarding early childhood language development milestones.

GPT-4 and NorBloom-7b's superior performance for certain age groups indicates model selection should be age-specific rather than universal across developmental stages. Prompt engineering strategies should account for the systematic bias toward linguistic overestimation, potentially incorporating explicit instructions to simplify language or use age-appropriate vocabulary.

However, purely computational metrics proved inadequate. Minimal differences in AoA scores, word frequency, and structural characteristics failed to capture qualitative differences human experts readily detected. While computational metrics remain valuable for large-scale analysis, human expert evaluation is essential for assessing authenticity in domains requiring nuanced developmental understanding. The 83\% bias toward early childhood vocabulary (under age 5) in our Norwegian database constrained computational differentiation between age groups, exemplifying resource limitations in low and medium resource languages. This highlights the need for more comprehensive developmental linguistic resources in languages other than high resource languages like English.

Our study has methodological constraints. The sample size was limited to two real interviews (one per age group) and ten LLM-generated texts (five models × two age groups), evaluated by ten education professionals. While our rigorous blind evaluation achieved strong inter-rater reliability, larger sample sizes including more real interviews across broader age ranges (e.g., 4-12 years) would strengthen findings and capture additional developmental stages.

\subsection{Implications and Future Directions}

While our study focused on Norwegian, the fundamental challenges identified are limited language specific training data, bias toward adult language patterns, and difficulty generating developmentally appropriate content—likely extend to other low or medium resource languages. Our methodological framework provides a template for systematic evaluation of LLM performance in similar contexts. Additionally, evaluation by education professionals suggests human oversight remains crucial in educational applications where developmental appropriateness directly impacts learning outcomes.

Future research should pursue four directions. First, developing of age-annotated child language corpora and automated metrics specifically designed for assessing developmental appropriateness. Second, conduct longitudinal studies tracking LLM-generated age-appropriate language effectiveness in actual educational applications, measuring learning outcomes and engagement. Third, expand to cross-linguistic studies comparing Norwegian findings with other low or medium resource languages to identify universal versus language-specific patterns. Fourth, explore advanced prompt engineering techniques that explicitly incorporate developmental linguistic knowledge, potentially using few-shot learning with authentic child language examples. Given the rapid evolution of language modeling technology, evaluation frameworks like ours require regular updating to assess newer architectures (e.g., GPT-5, Claude-4.5, Gemini) and training approaches. Our methodology provides a foundation for such ongoing assessment.

\section{Conclusion}

Our evaluation of five LLMs' ability to generate age-appropriate Norwegian child-like conversations reveals both promising capabilities and significant limitations in current approaches. While some models, particularly GPT-4 and NorBloom-7b, demonstrate reasonable capability in approximating child-like language patterns, substantial challenges remain in achieving authentic developmental appropriateness, especially for older children. The finding that education professionals achieved strong inter-rater reliability while showing significantly better accuracy for younger children validates both our evaluation methodology and the existence of distinct, recognizable patterns in child language development. However, the systematic tendency of most LLMs to generate language perceived as more linguistically advanced than target ages highlights fundamental gaps in current training paradigms that rely heavily on adult-authored content. Our work contributes a methodological framework for systematic evaluation of specialized language generation tasks while empirically demonstrating that model size alone does not guarantee better performance in domain-specific applications—smaller, specialized Norwegian models often outperformed larger general-purpose systems, suggesting that targeted adaptation strategies may be more effective than scaling approaches for low-resource language applications. A critical finding is the inadequacy of purely computational metrics for evaluating age-appropriate language generation, as objective measures failed to capture the qualitative differences human evaluators readily detected, underscoring the need for both comprehensive age-stratified linguistic resources and continued expert evaluation in specialized applications.

\section*{Ethics Statement}

This research received ethical approval from the Norwegian Agency for Shared Services in Education and Research (SIKT, project number 385370), titled "Child Speech Analysis." Our methodology for collecting and processing data from children aged 4-16 years followed strict ethical protocols to ensure participant privacy and data security.

Data collection involved face-to-face structured interviews conducted between an adult researcher and children, focusing exclusively on general, non-personal topics such as favorite foods, toys, animals, and everyday activities. All interviews were audio-recorded during sessions to ensure accurate data capture. Immediately following each interview, recordings were transcribed verbatim, then fully anonymized to remove all identifying information, including names, locations, and any personal details. After transcription and anonymization were completed, all original audio recordings were permanently deleted to minimize privacy risks and ensure data security.

Participation in the study was voluntary, with informed consent obtained from parents/guardians and age-appropriate assent from children. The ten education professionals who served as evaluators for the blind assessment phase were compensated with gift cards valued at 200 Norwegian Kroner for their time and expertise. All procedures were designed to safeguard participant well-being, with particular attention to creating comfortable, non-threatening interview environments appropriate for young children.


\bibliographystyle{acl_natbib}
\bibliography{references}

\clearpage
\appendix
\section{Child Lexical Study Data}
\label{app:child-lexical}

This appendix details the 12 text samples evaluated by education professionals in our user study, presented in the order they appeared in the evaluation form.

\begin{table}[h]
\footnotesize
\centering
\caption{Text samples used in the user study evaluation, listed in the order presented to evaluators.}
\label{app:sample-details}
\begin{tabular}{clcc}
\hline
\textbf{Study Order} & \textbf{Source} & \textbf{Target Age} & \textbf{Gender Persona} \\ 
\hline
1 & Real interview & 9 years & Boy \\ 
2 & RUTER-LLAMA-2-13b & 5 years & Boy \\
3 & GPTSW & 5 years & Boy \\ 
4 & RUTER-LLAMA-2-13b & 9 years & Girl \\
5 & GPT-4 & 5 years & Boy \\ 
6 & GPT-4 & 9 years & Girl \\ 
7 & NorMistral-7b & 5 years & Boy \\
8 & NorBloom-7b & 9 years & Girl \\ 
9 & Real interview & 5 years & Girl \\ 
10 & GPTSW & 9 years & Girl \\ 
11 & NorMistral-7b & 9 years & Girl \\
12 & NorBloom-7b & 5 years & Boy \\ 
\hline
\end{tabular}
\footnotesize
\end{table}

\clearpage


\section*{Supplementary material}

# Aldersvurdering av språk for yngre aldre.

Velkommen til vår studie om **aldersvurdering av generert tale** for barn i yngre aldre. Her ønsker vi å evaluere hvordan datamaskiner kan generere samtaler som etterligner talen og språket til unge individer ved å bruke AI og maskinlæring, eller nærmere bestemt ulike språkmodeller. Slike språkmodeller er designet for å forstå og generere naturlig språk som ligner menneskelig tekst og tale. Din ekspertise er viktig for å hjelpe oss med å vurdere alderen til barnet og kvaliteten til den datamaskin-genererte teksten.

**Oppgavebeskrivelse**:
Du vil bli presentert for flere tekstprøver hvor talen til barnet er omgjort til tekst (såkalt transkribert tekst fra lyd). Disse prøvene er generert av forskjellige typer språkmodeller og er ment å representere forskjellige samtaler med barn basert på barnets alder. Med andre ord, vi har bedt datamaskinen simulere et barn med alder X når den svarer på spørsmål vi stiller, og din oppgave er å lese hver tekstprøve nøye og deretter prøve å anslå alderen til barnet basert på de språklige egenskapene, innholdet, og modenheten i teksten. Du skal også begrunne svaret ditt.

Vennligst merk at tekstene er generert av en maskin (ikke ekte) og at de da ikke er basert på faktisk tale fra barn i en spesifikk alder. Merk også at siden det er flere språkmodeller gjentas de samme spørsmålene flere ganger.

* Indicates required question

**Bakgrunnsinformasjon**

1. Yrke *

   _____

2. Hva slags forhold har du til barn? *

   *Check all that apply.*

   ☐ Jeg er forelder.
   ☐ Jeg jobber i barnehage.
   ☐ Jeg jobber på skole.
   ☐ Jeg forsker på barn.
   ☐ Annet

3. Hvor mange år med erfaring har du med å jobbe med barn? *

   *Mark only one oval.*

   ○ 0-1 år
   ○ 1-3 år
   ○ 3-5 år
   ○ 5-7 år
   ○ 7-10 år
   ○ 10 eller flere år

4. Hvilken aldersgruppe av barn har/hadde du å gjøre med? *

   _____

5. Har du tidligere deltatt i studier relatert til språkutvikling hos barn? *

   *Mark only one oval.*

   ○ Ja
   ○ Nei

6. Hvis ja, vennligst beskriv kort din rolle og studiens natur.

   _____

   *Skip to question 7*

**Aldersvurdering av Tekstprøver - Test 1**

Nå skal du lese en samtale mellom et barn og en intervjuer. Datamaskinen simulerer da barnet med en viss alder. Prøv å anslå alderen på barnet.

For å fjerne åpenbare hint om alder så har vi erstattet "**alder**" i spørsmålet med "**##**", og fraser som "**i barnehagen**" og "**på skolen**" er erstattet med "**i ********". Eventuelle "**navn**" eller "**identifikasjonsopplysninger**" i svar erstattes med "**x**" eller "**xx**".

**Intervju:**

*Spørsmål: Nå er du ## år og da går du i ******. Og der har dere jo mange forskjellige fag. Fortell meg om ditt favorittfag i ******.*
Barn: Tja, jeg liker jo masse forskjellige ting jeg da. Matte, norsk kan bli litt kjedelig. Men eller liker jeg jo gym, spesielt slåball, turn, friidrett om sommern. Svømming.

*Spørsmål: Fortell meg hvorfor du liker dette/disse fagene så godt?*
Barn: Jeg liker de fordi det er morsomt. Gøy å gjøre forskjellige ting jeg liker.

*Spørsmål: Så vil jeg tro du driver med mye forskjellig utendørs i ******. Fortell meg hva du liker å gjøre da?*
Barn: Ja, det gjør vi. Det er mye leking i ******, der bygger vi hytte. Vi pleier å være en kompisgjeng der. Vi pleier å game endel, men ikke i ****** riktignok. Det får vi jo ikke lov til, selv om det er noen som har smuglet med seg telefonen i ******. Det gikk dårlig. Ellers er det jo å være ute da. Drive med zip-line, henge i apparater.

*Spørsmål: Og når du er hjemme etter ****** hva liker du å gjøre da?*
Barn: Være med kompiser, spille spill, av og til fotball. Basket ved skolegården der. Det liker jeg.

*Spørsmål: Drive du med noe form for sport eller andre aktiviteter etter ******? Hvis ja, fortell meg mer om det.*
Barn: Ishockey. Det er jo i hall vi driver med det. Av og til står vi på skøyter ute hvis det er bra is og kaldt nok. Ellers driver jeg egentlig ikke med noen andre sporter.

*Spørsmål: Har du noe favorittmat? Fortell meg om det.*
Barn: Ja, mye jeg liker. Men favoritten min er nok pizza og taco. Biffsnadder og pannekaker. Alt med smeltet ost er digg.

*Spørsmål: Det er ikke så lenge siden det var sommerferie. Fortell hva du gjorde i sommerferien?*
Barn: Vi har på x, sammen med andre familier. Var ikke veldig bra vær da, men vi badet mye i bassenget. Ellers var vi på fjelltur og i x, sammen med X og x, og x. Vi var ikke i utlandet i år, det skal vi neste år. I Påsken.

*Spørsmål: Fortell meg mer om sommerferien.*
Barn: Det var ganske kult å være på x da. Det var mye å finne på der. Bra fotballbane og bra strand, hvor vi var nesten hver dag.

*Spørsmål: Og når det er vinter, hva liker du gjøre da?*
Barn: Ishockey inne og skøyter ute. Blir mye av det, mye treninger. Jeg liker slalom også, snowboard har jeg prøvd. Det var mye vanskeligere enn jeg trodde da. Datt hele tids liksom.

*Spørsmål: Jeg antar du har spist is i sommer også. Hva er din favorittis?*
Barn: Ja, is har jeg spist. På brygga, ofte om kvelden. Softis med sjokoladestrø. Kuleis i sånn beger, forskjellige typer smaker. Liker best med sjokolade egentlig. Alt med sjokolade.

7. Etter din erfaring, hva er den beregnede alderen til barnet?(Test 1) *

   _____

8. Hva er hovedgrunnen til de vurderingene du ga?(Test 1) *

   _____
   _____
   _____
   _____

   *Skip to question 9*

**Aldersvurdering av Tekstprøver - Test 2**

Nå skal du lese en samtale mellom et barn og en intervjuer. Datamaskinen simulerer da barnet med en viss alder. Prøv å anslå alderen på barnet.

For å fjerne åpenbare hint om alder så har vi erstattet "**alder**" i spørsmålet med "**##**", og fraser som "**i barnehagen**" og "**på skolen**" er erstattet med "**i ********". Eventuelle "**navn**" eller "**identifikasjonsopplysninger**" i svar erstattes med "**x**" eller "**xx**".

**Intervju:**

*Spørsmål: Nå vet jeg at du er ## år og at du går i ******. Hva er det du liker å gjøre der?*
Barn:   Jeg liker å leke med leker i ****** og spille med mine venner. Jeg liker også å gå på piknikker og leke med sand.

*Spørsmål: I ****** kan man jo leke både inne og ute. Fortell hva du liker å gjøre ute i ******?*
Barn:   Wooo! Jeg liker å leke ute i ******! Jeg liker å gå på en tur i skogen og finne de vakre fuglene. Jeg liker å spille på lekeplassen og klatre i trærne. Jeg liker også å leke i sandkassen og bygge sanden til forskjellige ting. Men det jeg liker best er å spille med ballene på den store gressplassen. Det er så mye moro!

*Spørsmål: Når dere er inne i ****** gjør dere kanskje mange ulike aktiviteter. Fortell hva du liker best å leke med når du er inne i ******?*
Barn:   Jeg liker å leke med lekeplassene og leke med mye ball! Jeg liker også å leke med lekekjøretøyene og leke med mine venner.

*Spørsmål: Hva liker du å leke med når du er hjemme? Fortell meg mer om det.*
Barn:   WOOHOO! Jeg liker å leke med LEGO-byggeriet og mine actionfigurer! Jeg har en helt egen LEGO-byggeplass og jeg kan bygge alle slags ting med dem. Jeg kan bygge hus, biler, og til og med fly! Jeg kan også legge dem til mine actionfigurer og leke med dem. Jeg har en superhelt som heter Batman og han er så kul! Jeg kan også leke med min datamaskin og spille spill. Jeg kan spille spill med mine venner og vi kan ha så mye moro.

*Spørsmål: Har du en favorittleke hjemme? Fortell meg mer om den.*
Barn:   Ja, jeg har en favorittleke hjemme! Den heter Teddy og er en liten, brun teddybjørn. Han er så myk og koselig, og jeg elsker å klappe ham og klemme ham når jeg er ute av humør. Han har en liten hjerte i brystet som lyser i mørket, og jeg kan alltid finne veien i rommet mitt når jeg ser den. Jeg har Teddy med meg når jeg går til ***** og alle mine venner liker ham også. Han er så snill og alltid klar til

*Spørsmål: Har du en favoritt rett du liker? Fortell meg mer om den.*
Barn:   Ja, jeg liker spaghetti! Det er den beste! Det er mye lett å spise og det er så morsomt å spise. Jeg liker å ha det med tomatsaus, men jeg liker også det med ost. Det er så morsomt å spise med en gaffel! Jeg vil gjerne ha det til middag hver kveld!

*Spørsmål: Det er ikke så lenge siden det var sommerferie. Fortell hva du gjorde i sommerferien?*
Barn:   Jeg gikk til stranden med familien min og spilte i sand og vann. Jeg lærte også å svømme og surfe. Jeg gikk på feriekolonien og hadde det så kul å leke med venner. Jeg gikk på turer og så på fugler og dyr i skogen. Jeg hadde det så kul i sommerferien!

*Spørsmål: Fortell meg mer om sommerferien.*
Barn:   Åh, det var så kul! Jeg gikk på ferie med familien min til stranden og hadde det så hyggelig med å bygge sandslott og leke i vannet. Jeg gikk også på en piknik i parken og spiste så mange søtsaker som jeg kunne. Jeg gikk også på camping med venner og vi hadde det så kul med å lage egen mat og bli utslitt på lekene. Jeg gikk også på et museum og så på mange interessante ting som jeg aldri hadde sett før. Og jeg lærte også no.

*Spørsmål: Og når det er vinter, hva liker du gjøre da?*
Barn:   Oooh! Da det er vinter, liker jeg å spille snøballkrig!

*Spørsmål: Jeg antar du har spist is i sommer også. Hva er din favorittis?*
Barn:   Ja, jeg elsker is! Min favoritt er rød fruktsmak. Jeg liker også vanilje og chokolade!

9. Etter din erfaring, hva er den beregnede alderen til barnet?(Test 2) *

    _____

10. Hva er hovedgrunnen til de vurderingene du ga?(Test 2) *

    _____
    _____
    _____
    _____

*Skip to question 11*

**Aldersvurdering av Tekstprøver - Test 3**

Nå skal du lese en samtale mellom et barn og en intervjuer. Datamaskinen simulerer da barnet med en viss alder. Prøv å anslå alderen på barnet.

For å fjerne åpenbare hint om alder så har vi erstattet "**alder**" i spørsmålet med "**##**", og fraser som "**i barnehagen**" og "**på skolen**" er erstattet med "**i ******". Eventuelle "**navn**" eller "**identifikasjonsopplysninger**" i svar erstattes med "**x**" eller "**xx**".

**Intervju:**

*Spørsmål: Nå vet jeg at du er ## år og at du går i ******. Hva er det du liker å gjøre i ******?*
Barn: Jeg liker å leke med vennene mine og ha det gøy. Vi kan leke med dukker eller leke ute. Jeg liker også å lære nye ting.

*Spørsmål: I ****** kan man jo leke både inne og ute. Fortell hva du liker å gjøre ute i ******?*
Barn: Jeg liker å leke med vann! Jeg elsker å plaske rundt i vannpytter, og jeg kan også bruke spade og rive til å gjøre morsomme figurer i jorden.

*Spørsmål: Når dere er inne i ****** gjør dere kanskje mange ulike aktiviteter. Fortell hva du liker best å leke med når du er inne i ******?*
Barn: Jeg liker best å leke med lekene mine, spesielt togbanen min. Jeg elsker å bygge broer, tunneler og veier til togene.

*Spørsmål: Hva liker du å leke med når du er hjemme? Fortell meg mer om det.*
Barn: Jeg liker å leke med dukkene mine og bilene mine. Jeg har to dukker og fem biler. Jeg liker å kle dem opp og leke at de er på eventyr sammen. En gang kledde jeg opp dukkene mine i fine kjoler og tok dem med ut for å spise på en fancy restaurant. Det var veldig gøy!

*Spørsmål: Har du en favorittleke hjemme? Fortell meg mer om den.*
Barn: Min favorittleke er en bil. Den har fire hjul og den kan kjøre veldig fort. Jeg liker å kjøre den rundt i huset og utenfor.

*Spørsmål: Har du en favoritt rett du liker? Fortell meg mer om den.*
Barn: Min favorittrett er macaroni og ost. Jeg elsker å spise det fordi det er så myk og kremaktig inni, og det har også en sprø skorpe på utsiden. Det er min favoritt fordi jeg kan spise det med en skje, og det er ikke for vanskelig.

*Spørsmål: Det er ikke så lenge siden det var sommerferie. Fortell hva du gjorde i sommerferien?*
Barn: Jeg var på ferie med mamma og pappa på hytta vår i Sverige. Vi dro dit hver sommer i mange år, og det var alltid så gøy! Vi badet i sjøen, fisket, spilte spill og koste oss med god mat og drikke.

*Spørsmål: Fortell meg mer om sommerferien.*
Barn: Sommerferien er en tid da folk tar en pause fra sine vanlige oppgaver og ansvar. Folk kan ta tid til å slappe av, leke og nyte naturen. De kan også reise til et annet sted, enten det er for ferie eller forretninger.

*Spørsmål: Og når det er vinter, hva liker du gjøre da?*
Barn: Jeg liker å bygge snømann og snøhytte.

*Spørsmål: Jeg antar du har spist is i sommer også. Hva er din favorittis?*
Barn: Sjokolade

11. Etter din erfaring, hva er den beregnede alderen til barnet?(Test 3) *

    _____

12. Hva er hovedgrunnen til de vurderingene du ga?(Test 3) *

    _____
    _____
    _____
    _____

*Skip to question 13*

**Aldersvurdering av Tekstprøver - Test 4**

Nå skal du lese en samtale mellom et barn og en intervjuer. Datamaskinen simulerer da barnet med en viss alder. Prøv å anslå alderen på barnet.

For å fjerne åpenbare hint om alder så har vi erstattet "**alder**" i spørsmålet med "**##**", og fraser som "**i barnehagen**" og "**på skolen**" er erstattet med "**i ******". Eventuelle "**navn**" eller "**identifikasjonsopplysninger**" i svar erstattes med "**x**" eller "**xx**".

## Intervju:

**Spørsmål:** *Nå er du ## og da går du i ******. Og der har dere jo mange forskjellige fag. Fortell meg om ditt favorittfag i ******.*
**Barn:** Oooh, det er virkelig vanskelig å velge et favorittfag, men jeg liker virkelig matematikk! Det er så interessant å lære om tall og regler og løsninger. Jeg elsker også å lære om historie og engelsk, men matematikk er definitivt mitt favorittfag.

**Spørsmål:** *Fortell meg hvorfor du liker dette/disse fagene så godt?*
**Barn:** Jeg liker matematikk siden jeg liker å løse problemer og finne svar. Det er som en puzzle som jeg må løse og jeg liker å se hvordan alle delene passer sammen. Jeg liker også å se hvordan matematikk brukes i hverdagen, som for eksempel når jeg løser oppgavene mine eller når jeg teller antall mennesker i rommet mitt.

**Spørsmål:** *Så vil jeg tro du driver med mye forskjellig utendørs i ******. Fortell meg hva du liker å gjøre da?*
**Barn:** Jeg liker å spille basketball og volleyball med venner i ******! Det er så morsomt å løpe rundt og være en del av laget. Jeg liker også å løpe rundt i ******s område og leke på gressplassen. Jeg elsker å være ute og få litt frisk luft!

**Spørsmål:** *Og når du er hjemme etter ****** hva liker du å gjøre da?*
**Barn:** Jeg liker å spille med leketøyene mine, lese bøker, se på filmer, spille med hunden min og spille videospill med vennene mine. Jeg liker også å hjelpe moren min med å lage mat eller gjøre noen håndverksprosjekter.

**Spørsmål:** *Drive du med noe form for sport eller andre aktiviteter etter ******? Hvis ja, fortell meg mer om det.*
**Barn:** Ja, jeg elsker å drive med sport etter ******! Jeg spiller basketball og går på treningssenteret. Jeg liker å trene og se resultatene etter hvert som jeg trener. Jeg liker også å spille med vennene mine og ha det gøy mens vi lærer nye ferdigheter. Det er alltid noe nytt å lære og det er så morsomt!

**Spørsmål:** *Har du noe favoritttmat? Fortell om det.*
**Barn:** Ja, jeg liker å spise pizza og iskrem. Jeg liker også å spise frukt, som epler og bananer.

**Spørsmål:** *Det er ikke så lenge siden det var sommerferie. Fortell hva du gjorde i sommerferien?*
**Barn:** Jeg var i sommerferie og gikk på stranden med familien min. Vi spilte i vannet og solen og lade oss lure i solen. Vi gikk på turer i naturen og så mange vakre stjerner om kvelden. Vi tilbrakte også mye tid med å lese bøker og spille spill. Jeg hadde det så flott og jeg kan ikke vente med å gjøre det igjen neste sommer!

**Spørsmål:** *Fortell meg mer om sommerferien.*
**Barn:** Jeg har hatt en flott sommerferie! Jeg har vært på ferie med familien min i en måned, og vi har hatt det veldig hyggelig. Vi har gått til stranden og badet i sjøen, spist iskrem og gått på utendørsaktiviteter som camping og sykling. Jeg har også vært på ferie med venner og vi har hatt det veldig hyggelig på en campingplass. Vi har hatt det så bra at jeg ikke vil at sommerferien skal ta slutt!

**Spørsmål:** *Og når det er vinter, hva liker du gjøre da?*
**Barn:** Ooh, spørsmål! Når det er vinter, liker jeg å gå på skøyter og spille snøballkamp med venner!

**Spørsmål:** *Jeg antar du har spist is i sommer også. Hva er din favorittis?*
**Barn:** Ja, jeg elsker å spise is i sommer! Min favorittis er vanilje. Jeg liker den så mye fordi den er søt og myk, og den smaker så godt. Jeg liker også å prøve forskjellige smaker, så jeg har prøvd noen andre isflavorer også, men vanilje er fortsatt min favoritt.

13. Etter din erfaring, hva er den beregnede alderen til barnet?(Test 4) *

    _____

14. Hva er hovedgrunnen til de vurderingene du ga?(Test 4) *

    _____
    _____
    _____
    _____

*Skip to question 15*

### Aldersvurdering av Tekstprøver - Test 5

Nå skal du lese en samtale mellom et barn og en intervjuer. Datamaskinen simulerer da barnet med en viss alder. Prøv å anslå alderen på barnet.

For å fjerne åpenbare hint om alder så har vi erstattet "**alder**" i spørsmålet med "**##**", og fraser som "**i barnehagen**" og "**på skolen**" er erstattet med "**i ********". Eventuelle "**navn**" eller "**identifikasjonsopplysninger**" i svar erstattes med "**x**" eller "**xx**".

## Intervju:

**Spørsmål:** *Nå vet jeg at du er ## år og at du går i ******. Hva er det du liker å gjøre i ******?*
**Barn:** Jeg liker veldig godt å leke ute i sandkassen og bygge slott. Jeg liker også å tegne og male, og så synes jeg det er veldig gøy når vi har sangleker. Men aller best liker jeg å leke med de andre barna i ******.

**Spørsmål:** *I ****** kan man jo leke både inne og ute. Fortell hva du liker å gjøre ute i ******?*
**Barn:** Ute i ****** liker jeg å leke i lekehuset og på klatrestativet. Jeg liker også å kjøre trøbiler og sykle på trehjulssykkel. Når det er vinter, er det gøy å ake og bygge snømann. Men det aller beste er nok å leke sisten med de andre barna.

**Spørsmål:** *Når dere er inne i ****** gjør dere kanskje mange ulike aktiviteter. Fortell hva du liker best å leke med når du er inne i ******?*
**Barn:** Inne i ****** liker jeg best å leke med togbanen og byggeklossene. Jeg liker også å tegne og male på tegnebordet, og å lese bøker i lesehjørnet. Noen ganger får vi lov til å leke med leire, og det synes jeg er veldig gøy. Men det aller beste er nok når vi leker rolleleker, da kan jeg late som jeg er en superhelt eller en ridder.

**Spørsmål:** *Hva liker du å leke med når du er hjemme? Fortell meg mer om det.*
**Barn:** Når jeg er hjemme, liker jeg å leke med Lego. Jeg liker å bygge forskjellige ting, som biler, hus og romskip. Jeg har også en bamse som jeg liker veldig godt, den heter Teddy. Noen ganger leker jeg også med små biler på et bilteppe jeg har. Men det aller beste jeg vet er når mamma eller pappa leker med meg, da kan vi finne på all slags gøy sammen.

**Spørsmål:** *Har du en favorittleke hjemme? Fortell meg mer om den.*
**Barn:** Ja, min favorittleke hjemme er Lego. Jeg har en stor kasse med mange forskjellige Lego-klosser. Jeg liker at jeg kan bygge nesten hva som helst med Lego. Noen ganger bygger jeg høye tårn, andre ganger bygger jeg romskip eller biler. Det er gøy å bruke fantasien og se hva jeg kan lage. Jeg liker også å følge med på instruksjonene og bygge settene som kommer med Lego. Det er alltid spennende å se hva det blir til slutt.

**Spørsmål:** *Har du en favoritt rett du liker? Fortell meg mer om den.*
**Barn:** Min favorittrett er pannekaker! Jeg elsker når mamma eller pappa lager pannekaker til middag. Jeg liker å ha syltetøy på, både jordbær- og blåbærsyltetøy. Noen ganger får jeg også litt sukker på. Det beste er når pannekakene er litt tykke og myke. Jeg liker også å hjelpe til med å røre i røra, det er gøy!

**Spørsmål:** *Det er ikke så lenge siden det var sommerferie. Fortell hva du gjorde i sommerferien?*
**Barn:** I sommerferien var jeg og familien min på hytta ved sjøen. Det var veldig godt å bade, så vi var mye på stranda. Vi bygget sandslott, og jeg lærte å svømme med armringer. Vi hadde også med oss en båt som vi kunne ro i, det var gøy! Noen dager dro vi på tur i skogen og grillet pølser. Og en dag dro vi til en dyrepark, der så jeg mange forskjellige dyr. Jeg likte best å se på apene og løvene. Det var en veldig gøy sommerferie!

**Spørsmål:** *Fortell meg mer om sommerferien.*
**Barn:** I sommerferien gikk vi også på fisketur. Pappa lærte meg å fiske og jeg fanget min første fisk! Det var en liten abbor. Jeg var veldig stolt. Vi hadde også piknik ved sjøen og jeg fikk lov til å være oppe sent og se på stjernene. Vi hadde også en stor grillfest med alle våre venner på hytta. Jeg fikk lov til å løpe rundt og leke med vennene mine hele dagen. Jeg husker også at vi hadde en regnværsdag hvor vi spilte brettspill og jeg lærte å spille Ludo. Det var en fantastisk sommerferie!

**Spørsmål:** *Og når det er vinter, hva liker du gjøre da?*
**Barn:** Når det er vinter, liker jeg veldig godt å være ute i snøen. Jeg liker å ake ned bakken på akebrettet mitt, og å bygge snømann. Noen ganger lager vi også snøhuler, det er veldig gøy. Jeg liker også å gå på ski. Pappa har lært meg å stå på ski, og selv om det er litt vanskelig, synes jeg det er gøy. Når vi kommer inn igjen, liker jeg å drikke varm kakao og spise pepperkaker. Jeg synes vinteren er en veldig morsom tid.

**Spørsmål:** *Jeg antar du har spist is i sommer også. Hva er din favorittis?*
**Barn:** Min favorittis er jordbæriskrem! Jeg liker den søte smaken av jordbær, og det er ekstra gøy når det er biter av ekte jordbær i. Noen ganger får jeg også en vaffelkjeks til å ha isen i, det synes jeg er veldig godt. Jeg liker også sjokoladeis, men jordbæriskrem er definitivt favoritten min.

15. Etter din erfaring, hva er den beregnede alderen til barnet? (Test 5) *

    _____

16. Hva er hovedgrunnen til de vurderingene du ga? (Test 5) *

    _____
    _____
    _____
    _____

*Skip to question 17*

### Aldersvurdering av Tekstprøver - Test 6

Nå skal du lese en samtale mellom et barn og en intervjuer. Datamaskinen simulerer da barnet med en viss alder. Prøv å anslå alderen på barnet.

For å fjerne åpenbare hint om alder så har vi erstattet "**alder**" i spørsmålet med "**##**", og fraser som "**i barnehagen**" og "**på skolen**" er erstattet med "**i ******** **". Eventuelle "**navn**" eller "**identifikasjonsopplysninger**" i svar erstattes med "**x**" eller "**xx**".

**Intervju:**

*Spørsmål: Nå er du ## og da går du i ******. Og der har dere jo mange forskjellige fag. Fortell meg om ditt favorittfag i ******.*
Barn: Mitt favorittfag i ****** er matematikk. Det er så gøy å løse problemer og puslespill. Jeg liker hvordan tallene passer sammen, og jeg synes det er spennende å lære nye ting i det faget. Og læreren min i matematikk er veldig snill og hjelper oss hvis vi ikke forstår noe. Jeg liker også at vi noen ganger spiller matematiske spill eller gjør aktiviteter som hjelper oss å forstå bedre

*Spørsmål: Fortell meg hvorfor du liker dette/disse fagene så godt?*
Barn: Jeg liker matematikk fordi det er som et stort puslespill. Hver gang jeg løser et problem, føler jeg meg veldig smart. Jeg liker også hvordan tallene har et bestemt mønster, og hvordan jeg kan bruke det jeg har lært til å løse nye problemer. Læreren min gjør også faget morsomt. Vi spiller spill og gjør aktiviteter som hjelper oss med å forstå konseptene bedre. Jeg liker å lære nye ting, og matematikk gir meg alltid noe nytt og spennende å tenke på.

*Spørsmål: Så vil jeg tro du driver med mye forskjellig utendørs i ******. Fortell meg hva du liker å gjøre da?*
Barn: Jeg liker å leke ute i friminuttene. Noen ganger spiller vi fotball eller andre ballspill. Jeg liker også å hoppe tau og klatre på klatrestativet. Vi har også en sandkasse der vi bygger sandslott. Når det er fint vær, tar lærerne oss ofte ut for å ha undervisning i friluft, noe som er veldig gøy. Jeg liker også å løpe rundt og leke fange med vennene mine.

*Spørsmål: Og når du er hjemme etter ****** hva liker du å gjøre da?*
Barn: Når jeg kommer hjem fra ******, liker jeg å slappe av litt først. Kanskje jeg ser på en tegneserie eller leser en bok. Etterpå gjør jeg leksene mine. Jeg liker også å tegne og male, det er veldig gøy. Noen ganger spiller jeg også spill med broren min. Og før jeg legger meg, leser mamma eller pappa en godnatthistorie for meg. Det er alltid koselig.

*Spørsmål: Drive du med noe form for sport eller andre aktiviteter etter ******? Hvis ja, fortell meg mer om det.*
Barn: Ja, jeg driver med ballett etter ******. Jeg har alltid likt å danse, så det er veldig gøy for meg. Vi lærer forskjellige ballettposisjoner og danser til forskjellige typer musikk. Jeg liker også kostymene vi får bruke under forestillingene. I tillegg til ballett, er jeg med i en bokklubb i ******. Vi leser forskjellige bøker og diskuterer dem sammen. Det er veldig spennende fordi jeg elsker å lese og det er gøy å høre hva de andre tenker om bøkene vi leser.

*Spørsmål: Har du noe favorittmat? Fortell meg om det.*
Barn: Ja, jeg elsker spaghetti bolognese! Det er favorittmaten min fordi jeg elsker smaken av kjøttsausen sammen med spaghettien. Og så liker jeg at mamma alltid drysser litt ost på toppen. Det er også en rett vi ofte lager hjemme når vi skal ha ekstra godt til middag. Jeg liker også å hjelpe til med å lage den. Det er gøy å røre i sausen og se hvordan alt blander seg sammen til en deilig rett.

*Spørsmål: Det er ikke så lenge siden det var sommerferie. Fortell hva du gjorde i sommerferien?*
Barn: I sommerferien dro vi på familietur til en hytte ved sjøen. Det var veldig gøy å bygge sandslott på stranden og svømme i havet. Vi gikk også på noen lange turer i skogen og plukket bær. Noen dager ble vi bare hjemme og hadde grilling i hagen og spilte spill sammen. Jeg fikk også besøke noen av vennene mine og vi hadde overnattingsfester. Det var en veldig morsom og avslappende sommerferie.

*Spørsmål: Fortell meg mer om sommerferien.*
Barn: Sommerferien var virkelig spesiell i år. Vi tilbrakte mye tid på hytta vår som ligger ved

sjøen. Det var så herlig å våkne opp til lyden av bølgene hver morgen. Vi fisket en del, og jeg lærte å filetere fisken selv. Det var litt ekkelt i begynnelsen, men etter hvert syntes jeg det var ganske gøy. Vi grillet fisken vi fanget, og den smakte fantastisk. Jeg elsker også å bade, så jeg brukte mye tid på å svømme og plaske i vannet. Noen ganger gikk vi på tur i skogen som ligger i nærheten av hytta. Der plukket vi bær som brukte til å lage deilige desserter. Jeg fikk også tid til å henge med vennene mine. Vi hadde overnattingsfester hvor vi så på filmer, spilte spill og lo mye. Det var trist da sommerferien var over, men jeg gleder meg til neste år.

*Spørsmål: Og når det er vinter, hva liker du gjøre da?*
Barn: Når det er vinter, liker jeg å leke ute i snøen. Jeg elsker å lage snømenn og snøfort. Noen ganger har vi snøballkriger med vennene mine. Jeg liker også å gå på ski med familien min, både langrenn og alpint. Det er så gøy å skli nedover bakkene! Når det er veldig kaldt ute, liker jeg å sitte inne med en kopp varm kakao og se på snøen som faller. Jeg liker også å lese bøker eller se filmer under et varmt teppe. Og selvfølgelig liker jeg juletiden veldig godt, med julepynt, julekaker og gaver.

*Spørsmål: Jeg antar du har spist is i sommer også. Hva er din favorittis?*
Barn: Ja, jeg elsker iskrem! Min favoritt er jordbær. Jeg elsker den søte og litt syrlige smaken av jordbær. Når det er veldig varmt ute, er det ingenting som er bedre enn en kald jordbæris. Jeg liker også sjokoladeis, men jordbær er definitivt min favoritt. Og noen ganger liker jeg å ha litt sjokoladesaus på toppen. Det er så deilig!

17. Etter din erfaring, hva er den beregnede alderen til barnet?(Test 6) *

_____

18. Hva er hovedgrunnen til de vurderingene du ga?(Test 6) *

_____
_____
_____
_____

*Skip to question 19*

**Aldersvurdering av Tekstprøver - Test 7**

Nå skal du lese en samtale mellom et barn og en intervjuer. Datamaskinen simulerer da barnet med en viss alder. Prøv å anslå alderen på barnet.

For å fjerne åpenbare hint om alder så har vi erstattet "**alder**" i spørsmålet med "**##**", og fraser som "**i barnehagen**" og "**på skolen**" er erstattet med "**i ******". Eventuelle "**navn**" eller "**identifikasjonsopplysninger**" i svar erstattes med "**x**" eller "**xx**".

**Intervju:**

*Spørsmål: Nå vet jeg at du er ## år og at du går i ******. Hva er det du liker å gjøre i ******?*
Barn: Jeg liker å leke med andre barn og å spille fotball.

*Spørsmål: I ****** kan man jo leke både inne og ute. Fortell hva du liker å gjøre ute i ******?*
Barn: Det er morsomt å leke med vann, og så kan man leke med vennene sine.

*Spørsmål: Når dere er inne i ****** gjør dere kanskje mange ulike aktiviteter. Fortell hva du liker best å leke med når du er inne i ******?*
Barn: Jeg liker å leke med dukker og biler.

*Spørsmål: Hva liker du å leke med når du er hjemme? Fortell meg mer om det.*
Barn: Jeg liker å leke med lego og dukker.

*Spørsmål: Har du en favorittleke hjemme? Fortell meg mer om den.*
Barn: Ja, jeg har en favorittleke hjemme. Den heter "Baby".

*Spørsmål: Har du en favoritt rett du liker? Fortell meg mer om den.*
Barn: Pannekaker med syltetøy.

*Spørsmål: Det er ikke så lenge siden det var sommerferie. Fortell hva du gjorde i sommerferien?*
Barn: Jeg var i ****** og lekte med de andre barna.

*Spørsmål: Fortell meg mer om sommerferien.*
Barn: Jeg har vært på ferie i Norge.

*Spørsmål: Og når det er vinter, hva liker du gjøre da?*
Barn: Jeg vil gå med votter.

*Spørsmål: Jeg antar du har spist is i sommer også. Hva er din favorittis?*
Barn: Jeg liker vaniljeis.

19. Etter din erfaring, hva er den beregnede alderen til barnet?(Test 7) *

_____

20. Hva er hovedgrunnen til de vurderingene du ga?(Test 7) *

_____
_____
_____
_____

*Skip to question 21*

**Aldersvurdering av Tekstprøver - Test 8**

Nå skal du lese en samtale mellom et barn og en intervjuer. Datamaskinen simulerer da barnet med en viss alder. Prøv å anslå alderen på barnet.

For å fjerne åpenbare hint om alder så har vi erstattet "**alder**" i spørsmålet med "**##**", og fraser som "**i barnehagen**" og "**på skolen**" er erstattet med "**i ******". Eventuelle "**navn**" eller "**identifikasjonsopplysninger**" i svar erstattes med "**x**" eller "**xx**".

**Intervju:**

*Spørsmål: Nå er du ## og da går du i ******. Og der har dere jo mange forskjellige fag. Fortell meg om ditt favorittfag i ******.*
Barn: Mitt favorittfag er engelsk. Det er fordi jeg kan snakke engelsk med vennene mine.

*Spørsmål: Fortell meg hvorfor du liker dette/disse fagene så godt?*
Barn: Jeg liker å lære om naturen.

*Spørsmål: Så vil jeg tro du driver med mye forskjellig utendørs i ******. Fortell meg hva du liker å gjøre da?*
Barn: Hopper opp og ned.

*Spørsmål: Og når du er hjemme etter ****** hva liker du å gjøre da?*
Barn: Jeg liker å være med venner, spille fotball, og se på TV.

*Spørsmål: Drive du med noe form for sport eller andre aktiviteter etter ******? Hvis ja, fortell meg mer om det.*
Barn: Jeg liker å spille volleyball og basketball.

*Spørsmål: Har du noe favorittmat? Fortell meg om det.*
Barn: Jeg liker alt. Jeg liker alt, men jeg liker ikke så godt fisk.

*Spørsmål: Det er ikke så lenge siden det var sommerferie. Fortell hva du gjorde i sommerferien?*
Barn: Jeg var på hytta og badet.

*Spørsmål: Fortell meg mer om sommerferien.*
Barn: Sommerferien varte fra juli til august. Vi reiste til Italia og Frankrike. Vi hadde en fin ferie.

*Spørsmål: Og når det er vinter, hva liker du gjøre da?*
Barn: Jeg liker å gå på ski.

*Spørsmål: Jeg antar du har spist is i sommer også. Hva er din favorittis?*
Barn: Jeg har ikke spist is i sommer, men jeg har spist is I fjor.

21. Etter din erfaring, hva er den beregnede alderen til barnet?(Test 8) *

23. Etter din erfaring, hva er den beregnede alderen til barnet?(Test 9) *

_____

24. Hva er hovedgrunnen til de vurderingene du ga?(Test 9) *

_____
_____
_____
_____
_____

Skip to question 25

**Aldersvurdering av Tekstprøver - Test 10**

Nå skal du lese en samtale mellom et barn og en intervjuer. Datamaskinen simulerer da barnet med en viss alder. Prøv å anslå alderen på barnet.

For å fjerne åpenbare hint om alder så har vi erstattet "**alder**" i spørsmålet med "**##**", og fraser som "**i barnehagen**" og "**på skolen**" er erstattet med "**i ********". Eventuelle "**navn**" eller "**identifikasjonsopplysninger**" i svar erstattes med "**x**" eller "**xx**".

26. Hva er hovedgrunnen til de vurderingene du ga?(Test 10) *

_____
_____
_____
_____

*Skip to question 27*

### Aldersvurdering av Tekstprøver - Test 11

Nå skal du lese en samtale mellom et barn og en intervjuer. Datamaskinen simulerer da barnet med en viss alder. Prøv å anslå alderen på barnet.

For å fjerne åpenbare hint om alder så har vi erstattet "**alder**" i spørsmålet med "**##**", og fraser som "**i barnehagen**" og "**på skolen**" er erstattet med "**i ******"**. Eventuelle "**navn**" eller "**identifikasjonsopplysninger**" i svar erstattes med "**x**" eller "**xx**".

**Intervju:**
*Spørsmål: Nå er du ## og da går du i ******. Og der har dere jo mange forskjellige fag. Fortell meg om ditt favorittfag i ******.*
Barn: Ja, jeg liker veldig godt matte. Og da har vi jo mange forskjellige typer matte, så jeg liker best den matten der vi kan regne i hodet.

*Spørsmål: Fortell meg hvorfor du liker dette/disse fagene så godt?*
Barn: Fordi jeg liker å lære og jeg er glad i å jobbe.

*Spørsmål: Så vil jeg tro du driver med mye forskjellig utendørs i ******. Fortell meg hva du liker å gjøre da?*
Barn: Jeg liker å være med venner, leke og ha det gøy.

*Spørsmål: Og når du er hjemme etter ****** hva liker du å gjøre da?*
Barn: Da liker jeg å tegne.

*Spørsmål: Drive du med noe form for sport eller andre aktiviteter etter ******? Hvis ja, fortell meg mer om det.*
Barn: Nei, jeg gjør ikke det.

*Spørsmål: Har du noe favorittmat? Fortell meg om det.*
Barn: Jeg liker ikke fisk, og jeg liker ikke kjøtt.

*Spørsmål: Det er ikke så lenge siden det var sommerferie. Fortell hva du gjorde i sommerferien?*
Barn: Jeg var i Hellas og der badet jeg. Jeg hadde på meg en blå badedrakt. Jeg var der med familien min.

*Spørsmål: Fortell meg mer om sommerferien.*
Barn: Jeg var på ferie i Hellas, jeg dro med familien min.

*Spørsmål: Og når det er vinter, hva liker du gjøre da?*
Barn: Da er jeg inne og ser på TV.

*Spørsmål: Jeg antar du har spist is i sommer også. Hva er din favorittis?*
Barn: Jeg liker vaniljeis best.

27. Etter din erfaring, hva er den beregnede alderen til barnet?(Test 11) *

_____

28. Hva er hovedgrunnen til de vurderingene du ga?(Test 11) *

_____
_____
_____
_____

*Skip to question 29*

### Aldersvurdering av Tekstprøver - Test 12

Nå skal du lese en samtale mellom et barn og en intervjuer. Datamaskinen simulerer da barnet med en viss alder. Prøv å anslå alderen på barnet.

For å fjerne åpenbare hint om alder så har vi erstattet "**alder**" i spørsmålet med "**##**", og fraser som "**i barnehagen**" og "**på skolen**" er erstattet med "**i ******"**. Eventuelle "**navn**" eller "**identifikasjonsopplysninger**" i svar erstattes med "**x**" eller "**xx**".

**Intervju:**
*Spørsmål: Nå vet jeg at du er ## år og at du går i ******. Hva er det du liker å gjøre i ******?*
Barn: Jeg liker å leke med lego og bygge med klosser.

*Spørsmål: I ****** kan man jo leke både inne og ute. Fortell hva du liker å gjøre ute i ******?*
Barn: Jeg liker å leke i sandkassen.

*Spørsmål: Når dere er inne i ****** gjør dere kanskje mange ulike aktiviteter. Fortell hva du liker best å leke med når du er inne i ******?*
Barn: Jeg liker å leke med biler.

*Spørsmål: Hva liker du å leke med når du er hjemme? Fortell meg mer om det.*
Barn: Jeg liker å leke med biler og lastebiler.

*Spørsmål: Har du en favorittleke hjemme? Fortell meg mer om den.*
Barn: Ja, jeg har en favorittleke. Den er laget av gummi og er stor og myk.

*Spørsmål: Har du en favoritt rett du liker? Fortell meg mer om den.*
Barn: Jeg liker å spise kylling og ris.

*Spørsmål: Det er ikke så lenge siden det var sommerferie. Fortell hva du gjorde i sommerferien?*
Barn: Jeg var i utlandet med familien min. Vi var i Hellas.

*Spørsmål: Fortell meg mer om sommerferien.*
Barn: Vi var på ferie til et sted som heter Toscana i Italia. Vi var på stranden, vi besøkte et slott, vi dro til en innsjø og vi var på et marked.

*Spørsmål: Og når det er vinter, hva liker du gjøre da?*
Barn: Jeg liker å ake.

*Spørsmål: Jeg antar du har spist is i sommer også. Hva er din favorittis?*
Barn: Jeg har spist is hele sommeren, men min favoritt er nok Ben & Jerry's.

29. Etter din erfaring, hva er den beregnede alderen til barnet?(Test 12) *

_____

30. Hva er hovedgrunnen til de vurderingene du ga?(Test 12) *

_____
_____
_____
_____



\end{document}